\documentclass[preprint,authoryear,11pt]{elsarticle}

\usepackage{graphicx}
\usepackage{amssymb}
\usepackage{amsthm}
\usepackage[cmex10]{amsmath}
\usepackage{url}
\usepackage{lineno}

\journal{Knowledge-based Systems}

\begin{document}

\begin{frontmatter}

\title{An Empirical Evaluation of Similarity Measures\\ for Time Series Classification}

\author{Joan Serr\`a}
\author{Josep Ll.\ Arcos}
\address{Artificial Intelligence Research Institute (IIIA-CSIC),\\ Spanish National Research Council,\\ 08193 Bellaterra, Barcelona, Spain.}

\begin{abstract}
Time series are ubiquitous, and a measure to assess their similarity is a core part of many computational systems. In particular, the similarity measure is the most essential ingredient of time series clustering and classification systems. Because of this importance, countless approaches to estimate time series similarity have been proposed. However, there is a lack of comparative studies using empirical, rigorous, quantitative, and large-scale assessment strategies. In this article, we provide an extensive evaluation of similarity measures for time series classification following the aforementioned principles. We consider 7 different measures coming from alternative measure `families', and 45 publicly-available time series data sets coming from a wide variety of scientific domains. We focus on out-of-sample classification accuracy, but in-sample accuracies and parameter choices are also discussed. Our work is based on rigorous evaluation methodologies and includes the use of powerful statistical significance tests to derive meaningful conclusions. The obtained results show the equivalence, in terms of accuracy, of a number of measures, but with one single candidate outperforming the rest. Such findings, together with the followed methodology, invite researchers on the field to adopt a more consistent evaluation criteria and a more informed decision regarding the baseline measures to which new developments should be compared.
\end{abstract}


\end{frontmatter}
\linenumbers

\section{Introduction}
\label{sec:Intro}

Data in the form of time series pervades a large number of scientific domains~\citep{Keogh11AAAI,Keogh11DATA}. Observations that unfold over time usually represent valuable information subject to analysis, classification, indexing, prediction, or interpretation~\citep{Kantz04BOOK,Han05BOOK,Liao05PR,Fu11EAAI}. Real-world examples include financial data (e.g.,~stock market fluctuations), medical data (e.g.,~electrocardiograms), computer data (e.g.,~log sequences), or motion data (e.g,.~location of moving objects). Even object shapes or handwriting can be effectively transformed into time series, facilitating their analysis and retrieval~\citep{Keogh11DATA,Keogh09VLDB}.

A core issue when dealing with time series is determining their pairwise similarity, i.e.,~the degree to which a given time series resembles another. In fact, a time series similarity (or dissimilarity) measure is central to many mining, retrieval, clustering, and classification tasks~\citep{Han05BOOK,Liao05PR,Fu11EAAI,Keogh03DMKD}. Furthermore, there is evidence that simple approaches to such tasks exploiting generic time series similarity measures usually outperform more elaborate, sometimes specifically-targeted strategies. This is the case, for instance, with time series classification, where a one-nearest neighbor approach using a well-known time series similarity measure was found to outperform an exhaustive list of alternatives~\citep{Xi06ICML}, including decision trees, multi-scale histograms, multi-layer perceptron neural networks, order logic rules with boosting, or multiple classifier systems.

Deriving a measure that correctly reflects time series similarities is not straightforward. Apart from dealing with high dimensionality~\citep[time series can be roughly considered as multi-dimensional data;][]{Han05BOOK}, the calculation of such measures needs to be fast and efficient~\citep{Keogh03DMKD}. Indeed, with better information gathering tools, the size of time series data sets may continue to increase in the future. Moreover, there is the need for generic/multi-purpose similarity measures, so that they can be readily applied to any data set, whether this application is the final goal or just an initial approach to a given task. This last aspect highlights another desirable quality for time series similarity measures: their robustness to different types of data~\citep[cf.][]{Keogh03DMKD,Wang12DMKD}.

Over the years, several time series similarity measures have been proposed~\citep[for pointers to such measures see, e.g.,][]{Liao05PR,Fu11EAAI,Wang12DMKD}. Nevertheless, few quantitative comparisons have been made in order to evaluate their efficacy in a multiple-data framework~\citep{Keogh03DMKD}. Apart from being an interesting and important task by itself~\citep{Keogh11AAAI}, and as opposed to clustering~\citep{Liao05PR}, time series classification offers the possibility to straightforwardly assess the merit of time series similarity measures under a controlled, objective, and quantitative framework.

In a recent study,~\citet{Wang12DMKD} perform an extensive comparison of classification accuracies for 9 measures (plus 4 variants) across 38 data sets coming from various scientific domains. One of the main conclusions of the study is that, even though the newly proposed measures can be theoretically attractive, the efficacy of some common and well-established measures is, in the vast majority of cases, very difficult to beat. Specifically, dynamic time warping~\citep[DTW;][]{Berndt94KDD} is found to be consistently superior to the other studied measures (or, at worst, for a few data sets, equivalent). In addition, the authors emphasize that the Euclidean distance remains a quite accurate, robust, simple, and efficient way of measuring the similarity between two time series. Finally, by looking in detail at the results presented by~\citet{Wang12DMKD}, we can spot a group of time series similarity measures that seems to have an efficacy comparable to DTW: those based on edit distances. In particular, the edit distance for real sequences~\citep[EDR;][]{Chen05SIGMOD} seems to be very competitive, if not slightly better than DTW. Interestingly, none of the three measures above was initially targeted to generic time series data, but were introduced with hindsight~\citep{Agrawal93FDOA,Berndt94KDD,Chen05SIGMOD}. The intuition behind Euclidean distance relates to spatial proximity, DTW was initially devised for the specific task of spoken word recognition~\citep{Sakoe78TASLP}, and edit distances were introduced for measuring the dissimilarity between two strings~\citep{Levenshtein66SPD}.

The study by~\citet{Wang12DMKD} is, to the best of our knowledge, the only comparative study dealing with time series classification using multiple similarity measures and a large collection of data. In general, the studies introducing a new measure only compare against a few other measures\footnote{In the majority of cases, as our results will show, not the most appropriate ones.}, and usually using a reduced data set corpus~\citep[cf.][]{Keogh03DMKD}. Furthermore, there is a lack of agreement in the literature regarding evaluation methodologies. Besides, statistical significance is usually not studied or, at best, improperly evaluated. This is very inconvenient, as robust evaluation methodologies and statistical significance are the principal tools by which we can establish, in a formal and rigorous way, differences across the considered measures~\citep{Salzberg97DMKD,Hollander99BOOK,Demsar07JMLR}. In addition, the optimal parameter values for every measure are rarely discussed. All these issues impact the scientific development of the field as one is never sure, e.g.,~of which measure should be used as a baseline for future developments, or of which parameters are the most sensible choice.

In this work, we perform an empirical evaluation of similarity measures for time series classification. We follow the initiative by~\citet{Wang12DMKD}, and consider a big pool of publicly-available time series data sets (45 in our case). However, instead of additionally focusing on representation methods, computational/storage demands, or more theoretical issues, we here take a pragmatic approach and restrict ourselves to classification accuracy. We believe that this is the most important aspect to be considered in a first stage and that, in contrast to the other aforementioned issues, it is not sufficiently well-covered in the existing literature. As for the considered measures, we decide to include DTW and EDR, as these were found to generally achieve the highest accuracies among all measures compared in~\citet{Wang12DMKD}. Apart from these two, we choose the Euclidean distance plus 4 different measures not considered in such study, making up to a total of 7. 
Further important contributions that differentiate the current work from previous studies include (a)~an extensive summary and background of the considered measures, with basic formulations, applications, and references, (b)~the formalization of a robust evaluation methodology, exploiting standard out-of-sample cross-validation strategies, (c)~the use of rigorous statistical significance tests in order to assess the superiority of a given measure, (d)~the evaluation of both train and test accuracies, and (e) the assessment of the optimal parameters for each measure and data set.

The rest of the paper is organized as follows. Firstly, we provide the background on time series similarity measures, outline some of their applications, and detail their calculation (Sec.~\ref{sec:Measures}). Next, we explain the proposed evaluation methodology (Sec.~\ref{sec:Eval}). Subsequently, we report the obtained results (Sec.~\ref{sec:Results}). A conclusion section ends the paper (Sec.~\ref{sec:Conclu}).

\section{Time series similarity measures}
\label{sec:Measures}

The list of approaches for dealing with time series similarity is vast, and a comprehensive enumeration of them all is beyond the scope of the present work~\citep[for that, the interested reader is referred to][]{Gusfield97BOOK,Wang12DMKD,Han05BOOK,Liao05PR,Marteau09TPAMI,Fu11EAAI}. In this section, we present several representative examples of different `families' of time series similarity measures~\citep{Liao05PR,Wang12DMKD}: lock-step measures (Euclidean distance), feature-based measures (Fourier coefficients), model-based measures (auto-regressive), and elastic measures (DTW, EDR, TWED, and MJC). An effort has been made in selecting the most standard measures of each group, emphasizing the approaches that are reported to have good performance. We also try to avoid measures with too many parameters, since such parameters may be difficult to learn in small training data sets and, furthermore, could lead to over-fitting. Alternative measures found to be consistently less accurate than DTW or EDR are not considered~\citep{Wang12DMKD}. Apart from all the aforementioned measures, we also include a random measure, consisting of a uniformly distributed random number between 0 and 1. This will act as our random baseline.

\subsection{Euclidean distance}
\label{sec:Euclidean}

\begin{figure}[!t]
	\begin{center}
	\includegraphics{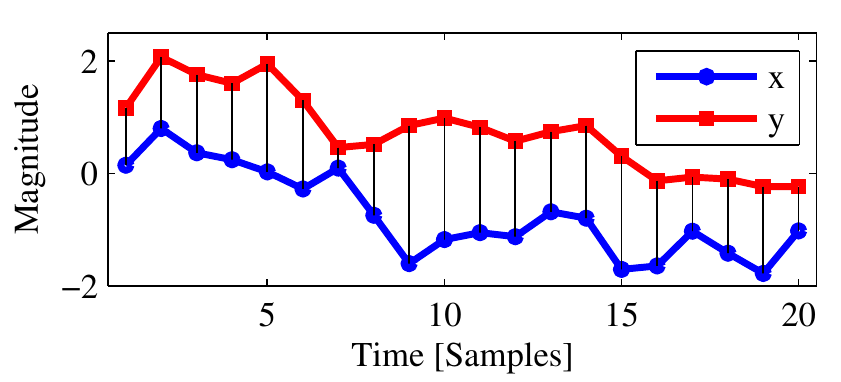}\\
	\includegraphics{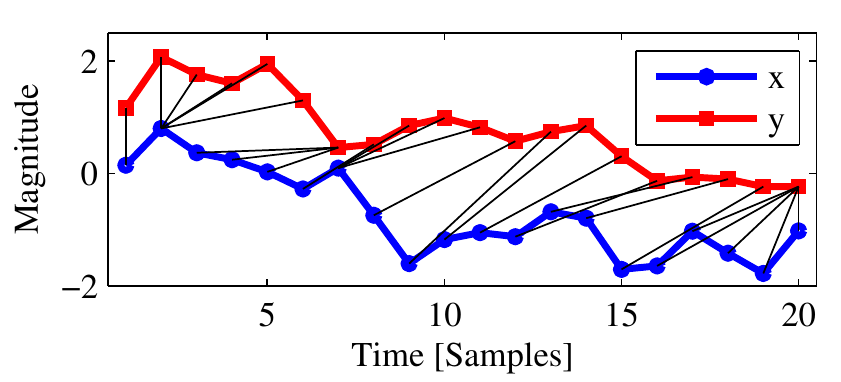}\\
	\includegraphics{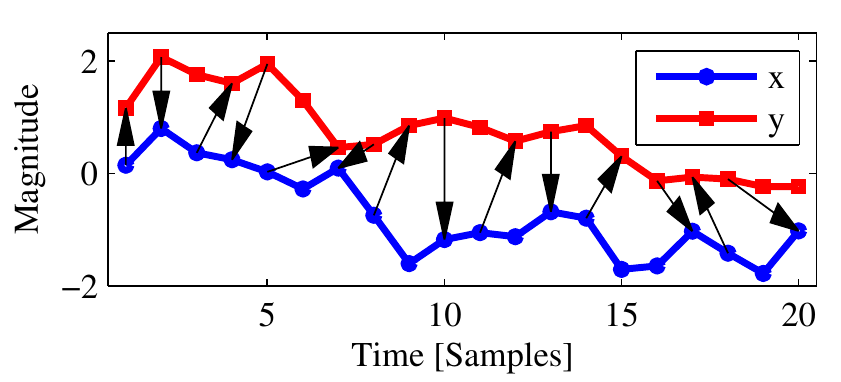}
	\end{center}
	\caption{Examples of dissimilarity calculations between time series $\textbf{x}$ and $\textbf{y}$: Euclidean distance (top), DTW alignment (center), and MJC (bottom). See text for details.}
	\label{fig:Examples}
\end{figure}

The simplest way to estimate the dissimilarity between two time series is to use any L$_n$ norm such that
\begin{equation}
	d_{\text{L}_n}(\textbf{x},\textbf{y}) = \left( \sum_{i=1}^M (x_i - y_i)^n \right)^{\frac{1}{n}} ,
	\label{eq:Euc}
\end{equation}
where $n$ is a positive integer, $M$ is the length of the time series, and $x_i$ and $y_i$ are the $i$-th element of time series $\textbf{x}$ and $\textbf{y}$, respectively. Measures based on L$_n$ norms correspond to the group of so-called lock-step measures~\citep{Wang12DMKD}, which compare samples that are at exactly the same temporal location (Fig.~\ref{fig:Examples}, top). Notice that in case the time series $\textbf{x}$ and $\textbf{y}$ not being of the same length, one can always re-sample one to the length of the other, an approach that works well for a number of data sources~\citep{Keogh03DMKD}. 

Using Eq.~\ref{eq:Euc} with $n=2$ we obtain the Euclidean distance, one of the most used time series dissimilarity measures, favored by its computational simplicity and indexing capabilities. Applications range from early classification of time series~\citep{Xing11KAIS} to rule discovery in economic, communications, and ecological time series~\citep{Das98KDDM}. Some authors state that the accuracy of the Euclidean distance can be very difficult to beat, specially for large data sets containing many time series~\citep[cf.][]{Wang12DMKD}. To the best of our knowledge, these claims are only quantitatively supported by one-nearest neighbor classification experiments using two artificially-generated/synthetic data sets~\citep{Geurts02THESIS}. We believe that such claims need to be carefully assessed with extensive experiments and under broader conditions, considering multiple measures, different distance-exploiting algorithms, and real-world data sets.

\subsection{Fourier coefficients}
\label{sec:FC}

A simple extension of the Euclidean distance is not to compute it directly using the raw time series, but using features extracted from it. For instance, by first representing the time series by their Fourier coefficients (FC), one uses
\begin{equation}
	d_{\text{FC}}(\textbf{x},\textbf{y}) = \left( \sum_{i=1}^{\theta} (\hat{x}_i - \hat{y}_i)^2 \right)^{\frac{1}{2}} ,
	\label{eq:FCDist}
\end{equation}
where $\hat{x}_i$ and $\hat{y}_i$ are complex value pairs denoting the $i$-th Fourier coefficient of $\hat{\textbf{x}}$ and $\hat{\textbf{y}}$, the discrete Fourier transforms (DFT) of the raw time series~\citep{Oppenheim99BOOK}. Notice that in Eq.~\ref{eq:FCDist} we introduce the parameter $\theta$, the actual number of considered coefficients. Because of the symmetry of the DFT, the sum only needs to be performed, at most, over half of the coefficients, so that $\theta=M/2$. Notice that, by the Parseval theorem~\citep{Oppenheim99BOOK}, the Euclidean distance between FCs is equivalent to the standard Euclidean distance between the raw time series~\citep[see, e.g.,][]{Agrawal93FDOA}. However, having parameter $\theta$, one usually takes the opportunity to filter out high-frequency coefficients, i.e.,~coefficients $\hat{x}_i$ and $\hat{y}_i$ whose $i$ is close to $M/2$. This has the (sometimes desired) effect of removing rapidly-fluctuating components of the signal. Hence, if high frequencies are not relevant for the intended analysis or we have some high-frequency noise, this operation will usually carry some increase in accuracy. Furthermore, if $\theta$ is relatively small, similarity computations can be substantially accelerated.

Computing the Euclidean distance on a reduced set of features is an extremely common approach in literature. FCs are the standard choice for efficient time series retrieval, exploiting the aforementioned acceleration capabilities. Pioneering work includes Agrawal et al.~\citep{Agrawal93FDOA} and Faloutsos et al.~\citep{Faloutsos95SIGMOD} dealing with synthetic and financial data. More recent works use FCs with data from other domains. For instance, the case-based reasoning system of Montani et al.~\citep{Montani06AIM} uses FCs to compare medical time series. Apart from FCs, wavelet coefficients have been extensively used~\citep{Chan99ICDE}. For instance, Olsson et al.~\citep{Olsson04JIFS} use a wavelet analysis to remove noise and extract features in their system of fault diagnosis in industrial equipment. Research suggests that, although they provide some advantages, wavelet coefficients do not generally outperform FCs for the considered task~\citep{Wu00CIKM}. Comparatively less used time series features are based on singular value decomposition~\citep{Wu96ICKI}, piece-wise aggregate approximations~\citep{Keogh01KAIS}, or the coefficients of fitted polynomials~\citep{Cai04SIGMOD} among others.

\subsection{Auto-regressive models}
\label{sec:AR}

A further option for computing similarities between time series using features extracted from them is to employ time series models~\citep{Liao05PR,Fu11EAAI}. The main idea behind model-based measures is to learn a model of the two time series and then use its parameters for computing a similarity value. In the literature, several approaches follow this idea. For instance,~\citet{Maharaj00JC} uses the $p$-value of a chi-square statistic to cluster auto-regressive coefficients representing stationary time series. \citet{Ramoni02ML} present a Bayesian algorithm for clustering time series. They transform each series into a Markov chain and then cluster similar chains to discover the most probable set of generating processes. \citet{Povinelli04TKDE} use Gaussian mixture models of reconstructed phase spaces to classify time series of different sources. \citet{Serra12TASLP} study the use of the error of several learned models to identify similar time series corresponding to musical information.

In the present study we consider the use of auto-regressive (AR) models for time series feature extraction~\citep{Piccolo90JTSA}. Given an AR model of the form
\begin{equation}
	x_i = a_0 + \sum_{j=1}^{\eta} a_jx_{i-j} ,
	\label{eq:AR}
\end{equation}
where $a_j$ denotes the $j$-th regression coefficient and $\eta$ is the order of the model, we can estimate its coefficients, e.g.,~by the Yule-Walker function~\citep{Marple87BOOK}. Then, the dissimilarity between two time series can be calculated, for instance, using the Euclidean distance between their estimated coefficients~\citep{Piccolo90JTSA}, analogously as in Eq.~\ref{eq:FCDist}. The number of AR coefficients is controlled by the parameter $\eta$ which, similarly to $\theta$ with FCs, directly affects the final speed of similarity calculations (AR and FCs are usually estimated offline, prior to similarity calculations).

\subsection{Dynamic time warping}
\label{sec:DTW}

Dynamic time warping~\citep[DTW;][]{Sakoe78TASLP,Berndt94KDD} is a classic approach for computing the dissimilarity between two time series. It has been exploited in countless works: to construct decision trees~\citep{Rodriguez04SAC}, to retrieve similar shapes from large image databases~\citep{Bartolini05TPAMI}, to match incomplete time series in medical applications~\citep{Tormene09AIM}, to align signatures in an identity authentication task~\citep{Kholmatov05PRL}, etc. In addition, several extensions for speeding up its calculations exist~\citep{Keogh05KAIS,Salvador07IDA,Lemire09PR}.

DTW belongs to the group of so-called elastic dissimilarity measures~\citep{Wang12DMKD}, and works by optimally aligning (or `warping') the time series in the temporal domain so that the accumulated cost of this alignment is minimal (Fig.~\ref{fig:Examples}, center). In its canonical form, this accumulated cost can be obtained by dynamic programming, recursively applying
\begin{equation}
	D_{i,j} = f(x_i,y_j) + \text{min}\left\lbrace D_{i,j-1}, D_{i-1,j}, D_{i-1,j-1} \right\rbrace
	\label{eq:DTW}
\end{equation}
for $i=1,\dots,M$ and $j=1,\dots,N$, being $M$ and $N$ the lengths of time series $\textbf{x}$ and $\textbf{y}$, respectively. Except for the first cell, which is initialized to $D_{0,0}=0$, the matrix $D$ is initialized to $D_{i,j}=\infty$ for $i=0,1,\dots,M$ and $j=0,1,\dots,N$. In the case of dealing with uni-dimensional time series, the local cost function $f()$, also called sample dissimilarity function, is usually taken to be the square of the difference between $x_i$ and $y_j$~\citep{Berndt94KDD}, i.e.,~$f(x_i,y_j)=(x_i-y_j)^2$. In the case of dealing with multidimensional time series or having some domain-specific knowledge, the local cost function $f()$ must be chosen appropriately, although the Euclidean distance is often used. The final DTW dissimilarity measure typically corresponds to the total accumulated cost, i.e.,~$d_{\text{DTW}}(\textbf{x},\textbf{y})=D_{M,N}$. A normalization of $d_{\text{DTW}}$ can be performed on the basis of the alignment of the two time series, which is found by backtracking from $D_{M,N}$ to $D_{0,0}$~\citep{Rabiner93BOOK}. However, in preliminary analysis we found the normalized variant to be equivalent, or sensibly less accurate, than the unnormalized one.

The canonical form of DTW presented in Eq.~\ref{eq:DTW} can incorporate many variants. In particular, several constraints can be applied to the computation of $D$. A common constraint~\citep{Sakoe78TASLP} is to introduce a window parameter $\omega\in[0,N]$, such that the recursive formula of Eq.~\ref{eq:DTW} is only applied for $i=1,\dots,M$ and
\begin{equation}
	j=\text{max}\lbrace 1,i'-\omega \rbrace,\dots,\text{min}\lbrace N,i'+\omega \rbrace ,
	\label{eq:DTWConst}
\end{equation}
where $i'$ is progressively adjusted for dealing with different time series lengths, i.e.,~$i'=\lfloor iN/M \rceil$, using $\lfloor~\rceil$ as the round-to-the-nearest-integer operator. Notice that if $\omega=0$ and $N=M$, $d_{\text{DTW}}$ will correspond to the squared Euclidean distance (the value in $D_{M,N}$ will be the sum of the squared differences, see Eqs.~\ref{eq:Euc} and~\ref{eq:DTW}). Notice furthermore that, when $\omega=N$, we are using the unconstrained version of DTW (the constraints in Eq.~\ref{eq:DTWConst} have no effect). Thus, we include two DTW variants in a single formulation. In general, the introduction of constraints, and specially of the window parameter $\omega$, carries some advantages~\citep{Keogh03DMKD,Rabiner93BOOK,Wang12DMKD}. For instance, constraints prevent `pathological alignments' and, therefore, usually provide better similarity estimates (pathological alignments typically go beyond the main diagonal of $D$). Moreover, constraints allow for reduced computational costs, since only a percentage of the cells in $D$ needs to be examined~\citep{Sakoe78TASLP,Rabiner93BOOK}.

DTW currently stands as the main benchmark against which new similarity measures need to be compared~\citep{Xi06ICML,Wang12DMKD}. Very few measures have been proposed that systematically outperform DTW for a number of different data sources. These measures are usually more complex than DTW, sometimes requiring extensive tuning of one or more parameters. Additionally, it is often the case that no careful, rigorous, and extensive evaluation of the accuracy of such measures is done, and further studies fail to assess the statistical significance of their improvement. Thus we could say that the superiority of such measures is, at best, unclear. In this paper, we pay special attention to all these aspects in order to formally assess the considered measures under a common framework. As it will be shown, there exists a similarity measure outperforming DTW for a statistically significant margin (Sec.~\ref{sec:Results}).

\subsection{Edit distance on real sequences}
\label{sec:EDR}

Turning to previous evidence~\citep{Wang12DMKD}, we observe that perhaps the only measure able to seriously challenge DTW is the edit distance on real sequences~\citep[EDR;][]{Chen05SIGMOD}. The EDR corresponds to the extension of the original edit or Levensthein distance~\citep{Levenshtein66SPD} to real-valued time series. Such extensions are not commonplace, but recent research is starting to focus on them~\citep{Morse07SIGMOD,Marteau09TPAMI}. As noted by~\citet{Chen05SIGMOD}, EDR outperformed previous edit distance variants for time series similarity. 

The computation of the EDR can be formalized by a dynamic programming approach. Specifically, we compute
\begin{equation}
	D_{i,j} =
	\begin{cases}
		D_{i-1,j-1} \quad\quad\quad\quad\quad~\text{if } m(x_i,y_j)=1 \\
		1+\text{min}\left\lbrace D_{i,j-1}, D_{i-1,j}, D_{i-1,j-1} \right\rbrace \\
		\quad\quad\quad\quad\quad\quad\quad\quad\quad~\text{if } m(x_i,y_j)=0 ,
	\end{cases}
	\label{eq:EDR}
\end{equation}
for $i=1,\dots,M$ and $j=1,\dots,N$. The match function used is
\begin{equation}
	m(x_i,y_j) = \Theta\left(\varepsilon - f\left(x_i,y_j\right)\right) ,
	\label{eq:EDRMatch}
\end{equation}
where $\Theta()$ is the Heaviside step function such that $\Theta(z)=1$ if $z\geq 0$ and $0$ otherwise, and $\varepsilon\in[0,\infty)$ is a suitably chosen threshold parameter that controls the degree of resemblance between two time series samples being considered as a match. The first row of $D$ is initialized to $D_{i,0}=i$ for $i=0,1,\dots,M$ and the first column of $D$ to $D_{0,j}=j$ for $j=0,1,\dots,N$. Following~\citet{Chen05SIGMOD}, who initially reported some accuracy improvements of EDR over DTW, we set the local cost function $f()$ to the absolute difference between the sample values, i.e.,~$f(x_i,y_j)=\vert x_i-y_j\vert$. This has the additional advantage that we can easily relate $\varepsilon$ to the standard deviation of the time series (Sec.~\ref{sec:Eval_ParamChoice}).

\subsection{Time-warped edit distance}
\label{sec:TWED}

The time-warped edit distance~\citep[TWED;][]{Marteau09TPAMI} is perhaps the most interesting extension of dynamic programming algorithms like DTW and EDR. In a sense, it is a combination of these two. Like edit distances, TWED comprises a mismatch penalty $\lambda$ and, like dynamic time warping, it introduces a so-called stiffness parameter $\nu$, controlling its `elasticity'~\citep{Marteau09TPAMI}. For uniformly-sampled time series, the formulation of TWED corresponds to
\begin{equation}
	D_{i,j} = \text{min} \left\lbrace D_{i,j} + \Gamma_{\textbf{xy}}, D_{i-1,j} + \Gamma_{\textbf{x}}, D_{i,j-1} + \Gamma_{\textbf{y}} \right\rbrace ,
	\label{eq:TWED}
\end{equation}
for $i=1,\dots,M$ and $j=1,\dots,N$, with
\begin{equation}
\begin{array}{lcl}
\Gamma_{\textbf{xy}} & = & f(x_i,y_j)+f(x_{i-1},y_{j-1})+2\nu\vert i-j \vert ,\\
\Gamma_{\textbf{x}}  & = & f(x_i,x_{i-1})+\nu+\lambda ,\\
\Gamma_{\textbf{y}}  & = & f(y_j,y_{j-1})+\nu+\lambda ,\\
\end{array}
\end{equation}
where $f()$ can be any L$_\text{n}$ metric (Eq.~\ref{eq:Euc}). Following~\citet{Marteau09TPAMI}, and as done for EDR as well, we choose $f(x_i,y_j)=\vert x_i-y_j \vert$. Together with DTW and EDR, the final dissimilarity value is taken to be $d_{\text{TWED}}(\textbf{x},\textbf{y}) = D_{M,N}$.

An interesting aspect of TWED is that, in its original formulation~\citep{Marteau09TPAMI}, it takes time stamp differences into account. Therefore, it is able to cope with time series of different sampling rates, including down-sampled time series. A further interesting aspect, and contrasting to DTW and other measures, is that TWED is a metric~\citep{Marteau09TPAMI}. Thus, one can exploit the triangular inequality to speed up the search in the metric space. Finally, it is worth mentioning that the combination of the two previous characteristics results in a lower bound of the TWED dissimilarity, which can be used to speed up nearest neighbor retrieval.

\subsection{Minimum jump costs dissimilarity}
\label{sec:MJC}

The main idea behind the minimum jump costs dissimilarity measure~\citep[MJC;][]{Serra12ICCBR} is that, if a given time series $\textbf{x}$ resembles $\textbf{y}$, the cumulative cost of iteratively `jumping' between their samples should be small\footnote{An implementation of MJC is made available online by the authors: \url{http://www.iiia.csic.es/~jserra/downloads/2012_SerraArcos_MJC-Dissim.tar.gz} (last accessed on September 15, 2013).} (Fig.~\ref{fig:Examples}, bottom). Supposing that for the $i$-th jump we are at time step $t_x$ of time series $\textbf{x}$, and that we previously visited time step $t_y-1$ of $\textbf{y}$, the minimum jump cost is expressed as
\begin{equation}
	c^{(i)}_{\text{min}} = \text{min}\left\lbrace c_{t_x}^{t_y}, c_{t_x}^{t_y+1}, c_{t_x}^{t_y+2}, \dots \right\rbrace ,
	\label{eq:MinCost}
\end{equation}
where $c_{t_x}^{t_y+\Delta}$ is the cost of jumping from $x_{t_x}$ to $y_{t_y+\Delta}$ and $\Delta=0,1,2,\dots$ is an integer time step increment such that $t_y+\Delta\leq N$. After a jump is made, $t_x$ and $t_y$ are updated accordingly: $t_x$ becomes $t_x+1$ and $t_y$ becomes $t_y+\Delta+1$. In case we want to jump from $\textbf{y}$ to $\textbf{x}$, only $t_x$ and $t_y$ need to be swapped~\citep{Serra12ICCBR}.

To define a jump cost $c_{t_x}^{t_y+\Delta}$, the temporal and the magnitude dimensions of the time series are considered:
\begin{equation}
	c_{t_x}^{t_y+\Delta} = (\phi\Delta)^2 + f(x_{t_x},y_{t_y+\Delta}) ,
	\label{eq:Cost}
\end{equation}
where $\phi$ represents the cost of advancing in time and $f()$ is the local cost function, which we take to be $f(x_{t_x},y_{t_y+\Delta})=(x_{t_x}-y_{t_y+\Delta})^2$, similarly to what is done with DTW (Eq.~\ref{eq:DTW}). Notice that, akin to the general formulation of TWED, the term $(\phi\Delta)^2$ introduces a nonlinear penalty that depends on the temporal gap. Here, the value of $\phi$ is set proportional to the standard deviation $\sigma$ expected for the time series and, at the same time, proportional to the real-valued parameter $\beta\in[0,\infty)$, which controls how difficult is to advance in time~\citep[for more details see][]{Serra12ICCBR}. To obtain a symmetric dissimilarity measure, $d_{\text{MJC}}(\textbf{x},\textbf{y}) = \text{min} \left\lbrace d_{\text{XY}}, d_{\text{YX}} \right\rbrace$ can be used, where $d_{\text{XY}}$ and $d_{\text{YX}}$ are the cumulative MJCs obtained by starting at $x_1$ and $y_1$, respectively.

\section{Evaluation methodology}
\label{sec:Eval}

\subsection{Classification scheme}
\label{sec:Eval_ClasScheme}

The efficacy of a time series similarity measure is commonly evaluated by the classification accuracy it achieves~\citep{Keogh03DMKD,Wang12DMKD}. For that, the error ratio of a distance-based classifier is calculated for a given labeled data set, understanding the error ratio as the number of wrongly classified items divided by the total number of tested items. The standard choice for the classifier is the one-nearest neighbor (1NN) classifier. Following~\citet{Wang12DMKD}, we can enumerate several advantages of using this approach. First, the error of the 1NN classifier critically depends on the similarity measure used. Second, the 1NN classifier is parameter-free and easy to implement. Third, there are theoretical results relating the error of an 1NN classifier to errors obtained with other classification schemes. Fourth, some works suggest that the best results for time series classification come from simple nearest neighbor methods. For more details on these aspects we refer to \citet{Mitchell97BOOK,Hastie09BOOK}, and the references provided by~\citet{Wang12DMKD}.

\subsection{Data sets}
\label{sec:Eval_Data}

We perform experiments with 45 publicly-available time series data sets from the UCR time series repository~\citep{Keogh11DATA}. This is the world's biggest time series repository, and some authors estimate that it makes up to more than 90\% of all publicly-available, labeled data sets~\citep{Wang12DMKD}. The repository comprises synthetic, as well as real-world data sets, and also includes one-dimensional time series extracted from two-dimensional shapes~\citep{Keogh11DATA}. The 45 data sets considered here correspond to the totality of the UCR repository, as by March 2013. Within such data sets, the number of classes ranges from 2 to 50, the number of time series per data set ranges from 56 to 9,236, and time series lengths go from 24 to 1,882 samples. For further details on these data sets we refer to~\citep{Keogh11DATA}.

\subsection{Cross-validation}
\label{sec:Eval_CrossValid}

To properly assess a classifier's error, out-of-sample validation needs to be done~\citep{Salzberg97DMKD}. In our experiments, we follow a standard 3-fold cross-validation scheme using balanced data sets~\citep{Mitchell97BOOK,Hastie09BOOK}, i.e., using the same number of items per class. We repeat the validation 20 times and report average error ratios. Balancing the data sets allows for balanced error estimations regarding the class distribution, and repeating cross-fold validation several times allows for more precise estimations~\citep{Mitchell97BOOK,Hastie09BOOK}. The use of a cross-fold validation scheme is essential for avoiding the bias that a particular split of the data could introduce~\citep{Salzberg97DMKD,Hastie09BOOK}.

We also computed error ratios for the original splits provided in the UCR time series repository~\citep{Keogh11DATA}. This allowed us to confirm that the 1NN error ratios from our implementations of DTW and Euclidean distance agree with the values reported there. In addition, we observed that the error ratios obtained by such splits were substantially different from the ones obtained by cross-validation, up to the point of even modifying the ranking of some algorithms with respect to those error ratios in some data sets. This indicates a potential bias in such individual splits, an aspect that is well-known in the machine learning community~\citep{Salzberg97DMKD,Mitchell97BOOK,Hastie09BOOK}. We refer the interested reader to any machine learning textbook for a more in-depth discussion of cross-fold validation schemes and their appropriateness over individual splits. Besides, using a single split difficults statistical significance assessment (see below). A full account of the raw error ratios for all measures and data sets is available online\footnote{\url{http://www.iiia.csic.es/~jserra/downloads/2013_SerraArcos_AnEmpiricalEvaluation.tar.gz} (last accessed on September 15, 2013).}, including the error ratios for the aforementioned original splits.

\subsection{Statistical significance}
\label{sec:Eval_StatSig}

To assess the statistical significance of the difference between two error ratios we employ the well-known Wilcoxon signed-rank test~\citep{Hollander99BOOK}. The Wilcoxon signed-rank test is a non-parametric statistical hypothesis test used when comparing two repeated measurements (or related samples, or matched samples) in order to assess whether their population mean ranks differ. It is the natural alternative to the Student's $t$-test for dependent samples when the population distribution cannot be assumed to be normal~\citep{Hollander99BOOK}. For a given data set, we use as input the $20\times 3$ accuracy values obtained for each classifier (i.e.,~the test fold accuracies). Besides, for comparing similarity measures on a more global basis using all data sets, we employ as input the 45 average accuracy values obtained for each data set. Following common practice~\citep{Salzberg97DMKD,Hollander99BOOK}, the threshold significance level is set to 5\%. Additionally, to compensate for multiple pairwise comparisons, we apply the Holm-Bonferroni method~\citep{Holm79SJS}, a post-hoc statistical analysis method controlling the so-called family-wise error rate that is more powerful than the usual Bonferroni correction~\citep{Demsar07JMLR}. 

\subsection{Parameter choices}
\label{sec:Eval_ParamChoice}

Before performing the experiments, all time series from all data sets were z-normalized so that each individual time series had zero mean and unit variance. Furthermore, we optimized the measures' parameters in the training phase of our cross-validation. This optimization step consisted of a grid search within a suitable range of parameter values, forcing the same number of parameter combinations per algorithm (Table~\ref{tab:Params}). The values of the grid are chosen according to common practice and the specifications given in the papers introducing each measure (Sec.~\ref{sec:Measures}). Specifically, for FC we used 25 linearly-spaced integer values of $\theta\in[2,N/2]$. For AR we used 25 linearly-spaced integer values of $\eta\in[1,0.25N]$ (because of the z-normalization, we remove $a_0$ in Eq.~\ref{eq:AR}). For DTW we used 24 linearly-spaced integer values of $\omega\in[0,0.25N]$ plus $w=N$ (the unconstrained DTW variant). For EDR we used 25 linearly-spaced real values of $\varepsilon\in[0.02\sigma,\sigma]$, $\sigma$ being the standard deviation of the time series (because of the z-normalization $\sigma=1$). For TWED we used all possible 25 combinations for $\nu=[10^{-4},10^{-3},10^{-2},10^{-1},1]$ and $\lambda=[0,0.25,0.5,0.75,1]$. For MJC we used 24 linearly-spaced real values of $\beta\in[0,25]$ plus $\beta=10^{10}$ (in practice corresponding to the squared Euclidean distance variant, Eq.~\ref{eq:Cost}). After the grid search, the parameter value yielding to the lowest leave-one-out error ratio for the training set was kept for out-of-sample testing.

\begin{table*}[!t]
	\begin{center}
	\resizebox{1\linewidth}{!}{
	\begin{tabular}{lccccc}
	\hline
	\hline
	Measure	& Parameter	& Minimum value	& Maximum value	& Number of steps 	& Extra value\\
	\hline
	FC		& $\theta$	& $2	$			& $0.5N$			& 25				& - \\
	AR		& $\eta$		& $1$				& $0.25N$ 		& 25				& - \\
	DTW		& $\omega$	& $0$				& $0.25N$ 		& 24				& $N$ \\
	EDR		& $\varepsilon$ & $0.02\sigma$	& $\sigma$		& 25				& - \\
	TWED		& $\nu$		& $10^{-5}$		& $1$				& 5				& - \\
	TWED		& $\lambda$	& $0$				& $1$				& 5				& - \\
	MJC		& $\beta$	& $0$				& $25$				& 24				& $10^{10}$ \\
	\hline
	\hline
	\end{tabular}
	}
	\end{center}
	\caption{Parameter grid for the considered similarity measures (recall that $N$ corresponds to the length of the time series and, since we z-normalize all time series, $\sigma=1$). For DTW and MJC we consider an extra value corresponding to unconstrained DTW and to the Euclidean configuration of MJC, respectively. All parameter values were linearly spaced except $\nu$, which was logarithmically spaced.}
	\label{tab:Params}
\end{table*}

\section{Results}
\label{sec:Results}

\subsection{Classification performance: test}
\label{sec:Results_Test}

If we look at the overall results, we see that all considered measures clearly outperform the random baseline for practically all the 45 data sets (Table~\ref{tab:Errors}). Furthermore, we see that some of them achieve near-perfect accuracies for a number of data sets (e.g.,~\textit{CBF}, \textit{CinC\_ECG\_torso}, \textit{ECGFiveDays}, \textit{Two\_Patterns}, or \textit{TwoLeadECG}). However, no single measure achieves the best performance for all the data sets. The Euclidean distance is found to be the best-performing measure in 2 data sets, FC is the best-performing in 4 data sets, AR in 1, DTW in 6, EDR in 7, TWED in 20, and MJC in 5. If we count only the data sets where one measure statistically significantly outperforms the rest, the numbers reduce to 0 for Euclidean, 2 for FC, 1 for AR, 2 for DTW, 2 for EDR, 6 for TWED, and 0 for MJC. Thus, interestingly, there are some data sets where choosing a specific similarity measure can make a difference.

\begin{table*}[!th]
	\renewcommand{\tabcolsep}{0.26cm}
	\resizebox{1\linewidth}{!}{
	\begin{tabular}{rl|cccccccc}
	\hline
	\hline
\# & Data set  & Random  & Euc  & FC  & AR  & DTW  & EDR  & TWED  & MJC \\
	\hline
1 & \textit{50words}  &  0.969  &  0.503  &  0.685  &  0.867  &  0.332  &  0.289  &  \textbf{0.237}$^\ast$  &  0.319 \\
2 & \textit{Adiac}  &  0.970  &  0.345  &  \textbf{0.266}$^\ast$  &  0.725  &  0.355  &  0.423  &  0.335  &  0.346 \\
3 & \textit{Beef}  &  0.763  &  0.417  &  \textbf{0.390}  &  0.504  &  0.472  &  0.439  &  0.506  &  0.448 \\
4 & \textit{CBF}  &  0.655  &  0.013  &  0.358  &  0.432  &  0.000  &  0.002  &  \textbf{0.000}  &  0.001 \\
5 & \textit{ChlorineConcentration}  &  0.673  &  0.071  &  0.063  &  \textbf{0.038}$^\ast$  &  0.072  &  0.094  &  0.093  &  0.070 \\
6 & \textit{CinC\_ECG\_torso}  &  0.749  &  0.002  &  0.008  &  0.102  &  0.001  &  \textbf{0.000}$^\ast$  &  0.001  &  0.002 \\
7 & \textit{Coffee}  &  0.394  &  0.019  &  0.024  &  0.139  &  \textbf{0.014}  &  0.031  &  0.021  &  0.023 \\
8 & \textit{Cricket\_X}  &  0.913  &  0.378  &  0.348  &  0.713  &  0.209  &  0.237  &  \textbf{0.190}$^\ast$  &  0.253 \\
9 & \textit{Cricket\_Y}  &  0.928  &  0.423  &  0.411  &  0.814  &  0.222  &  0.224  &  \textbf{0.209}  &  0.267 \\
10 & \textit{Cricket\_Z}  &  0.920  &  0.380  &  0.353  &  0.731  &  0.212  &  0.235  &  \textbf{0.194}$^\ast$  &  0.254 \\
11 & \textit{DiatomSizeReduction}  &  0.744  &  0.008  &  0.011  &  0.222  &  0.010  &  0.016  &  0.012  &  \textbf{0.007} \\
12 & \textit{ECG200}  &  0.515  &  0.130  &  0.145  &  0.227  &  0.139  &  0.148  &  \textbf{0.109}  &  0.130 \\
13 & \textit{ECGFiveDays}  &  0.505  &  0.007  &  \textbf{0.000}  &  0.072  &  0.003  &  0.003  &  0.005  &  0.001 \\
14 & \textit{FaceAll}  &  0.931  &  0.139  &  0.152  &  0.649  &  0.053  &  0.019  &  \textbf{0.019}  &  0.034 \\
15 & \textit{FaceFour}  &  0.679  &  0.111  &  0.149  &  0.545  &  0.069  &  0.028  &  0.025  &  \textbf{0.024} \\
16 & \textit{FacesUCR}  &  0.929  &  0.138  &  0.148  &  0.648  &  0.052  &  0.019  &  \textbf{0.018}  &  0.041 \\
17 & \textit{Fish}  &  0.871  &  0.183  &  0.234  &  0.617  &  0.184  &  \textbf{0.084}  &  0.094  &  0.114 \\
18 & \textit{Gun\_Point}  &  0.506  &  0.058  &  0.031  &  0.149  &  0.023  &  \textbf{0.010}  &  0.017  &  0.014 \\
19 & \textit{Haptics}  &  0.793  &  0.604  &  0.610  &  0.678  &  0.554  &  0.611  &  \textbf{0.544}  &  0.563 \\
20 & \textit{InlineSkate}  &  0.862  &  0.524  &  0.601  &  0.497  &  0.462  &  0.456  &  0.416  &  \textbf{0.411} \\
21 & \textit{ItalyPowerDemand}  &  0.489  &  0.035  &  0.083  &  0.261  &  \textbf{0.033}  &  0.042  &  0.036  &  0.034 \\
22 & \textit{Lighting2}  &  0.488  &  0.297  &  0.281  &  0.450  &  0.162  &  0.220  &  \textbf{0.161}  &  0.254 \\
23 & \textit{Lighting7}  &  0.817  &  0.371  &  0.463  &  0.707  &  \textbf{0.252}  &  0.362  &  0.256  &  0.336 \\
24 & \textit{Mallat}  &  0.870  &  0.018  &  0.020  &  0.058  &  0.015  &  0.006  &  \textbf{0.006}  &  0.014 \\
25 & \textit{MedicalImages}  &  0.912  &  0.313  &  0.455  &  0.458  &  0.247  &  0.330  &  \textbf{0.228}  &  0.305 \\
26 & \textit{MoteStrain}  &  0.513  &  0.087  &  0.162  &  0.336  &  0.058  &  0.024  &  \textbf{0.021}  &  0.034 \\
27 & \textit{NonInvasiveFetalECG1}  &  0.978  &  0.171  &  0.213  &  0.401  &  0.175  &  0.186  &  0.182  &  \textbf{0.169} \\
28 & \textit{NonInvasiveFetalECG2}  &  0.975  &  \textbf{0.106}  &  0.146  &  0.296  &  0.107  &  0.118  &  0.108  &  0.110 \\
29 & \textit{OliveOil}  &  0.644  &  \textbf{0.104}  &  0.185  &  0.663  &  0.154  &  0.194  &  0.146  &  0.127 \\
30 & \textit{OSULeaf}  &  0.832  &  0.409  &  0.306  &  0.617  &  0.359  &  \textbf{0.191}$^\ast$  &  0.232  &  0.256 \\
31 & \textit{SonyAIBORobotSurface}  &  0.510  &  0.017  &  0.040  &  0.079  &  0.018  &  0.026  &  0.017  &  \textbf{0.015} \\
32 & \textit{SonyAIBORobotSurfaceII}  &  0.489  &  0.018  &  0.032  &  0.113  &  0.021  &  0.023  &  \textbf{0.016}  &  0.019 \\
33 & \textit{StarLightCurves}  &  0.671  &  0.124  &  \textbf{0.070}$^\ast$  &  0.274  &  0.083  &  0.107  &  0.097  &  0.109 \\
34 & \textit{SwedishLeaf}  &  0.932  &  0.196  &  0.142  &  0.376  &  0.129  &  0.101  &  \textbf{0.094}  &  0.100 \\
35 & \textit{Symbols}  &  0.838  &  0.038  &  0.074  &  0.260  &  0.019  &  \textbf{0.015}  &  0.016  &  0.018 \\
36 & \textit{Synthetic\_control}  &  0.834  &  0.087  &  0.393  &  0.511  &  \textbf{0.009}$^\ast$  &  0.047  &  0.014  &  0.034 \\
37 & \textit{Trace}  &  0.757  &  0.169  &  0.117  &  0.117  &  \textbf{0.000}$^\ast$  &  0.034  &  0.011  &  0.038 \\
38 & \textit{Two\_Patterns}  &  0.743  &  0.020  &  0.491  &  0.724  &  0.000  &  0.000  &  \textbf{0.000}  &  0.001 \\
39 & \textit{TwoLeadECG}  &  0.507  &  0.006  &  0.012  &  0.202  &  0.001  &  0.002  &  \textbf{0.001}  &  0.003 \\
40 & \textit{UWaveGestureLibrary\_X}  &  0.872  &  0.234  &  0.566  &  0.694  &  0.199  &  0.214  &  \textbf{0.192}$^\ast$  &  0.203 \\
41 & \textit{UWaveGestureLibrary\_Y}  &  0.876  &  0.288  &  0.631  &  0.645  &  \textbf{0.263}  &  0.280  &  0.265  &  0.267 \\
42 & \textit{UWaveGestureLibrary\_Z}  &  0.879  &  0.298  &  0.546  &  0.678  &  0.265  &  0.271  &  \textbf{0.250}$^\ast$  &  0.261 \\
43 & \textit{Wafer}  &  0.497  &  0.004  &  0.003  &  0.013  &  0.005  &  \textbf{0.002}  &  0.003  &  0.005 \\
44 & \textit{WordsSynonyms}  &  0.960  &  0.496  &  0.675  &  0.855  &  0.327  &  0.304  &  \textbf{0.251}$^\ast$  &  0.310 \\
45 & \textit{Yoga}  &  0.500  &  0.070  &  0.108  &  0.333  &  0.061  &  \textbf{0.034}  &  0.037  &  0.047 \\
	\hline
   & Average rank & 7.99 & 4.40 & 5.07 & 6.80 & 3.00 & 3.42 & \textbf{2.29} & 3.04 \\
	\hline
	\hline
	\end{tabular}
	}
	\caption{Error ratios for all considered measures and data sets. The symbol $^\ast$ denotes a statistically significant difference with respect to the other measures for a given data set ($p<0.05$, Sec.~\ref{sec:Eval_StatSig}). The last row contains the average rank of each measure across all data sets (i.e.,~the average position after sorting the errors for a given data set in ascending order).}
	\label{tab:Errors}
\end{table*}

Beyond accuracies, this latter aspect can potentially highlight inherent data set qualities. For instance, the fact that a feature/model-based measure clearly outperforms the others for a particular data set indicates that such time series may be very well characterized by the extracted features/fitted model (e.g.,~FC with \textit{Adiac} for features and AR with \textit{ChlorineConcentration} for models). In addition, the good or bad performance of Euclidean and elastic measures gives us an intuition of the importance of alignments, warping, or sample correspondences (e.g.,~these may be very important for \textit{Trace} and the three \textit{Face} data sets, where there is an order of magnitude difference between Euclidean and warping-based measures, but not much for \textit{DiatomSizeReduction} or \textit{NonInvasiveFetalECG2}, where Euclidean gets numbers that are very close, or even better than the ones obtained by the warping-based measures).

\begin{figure}[t]
	\begin{center}
	\includegraphics{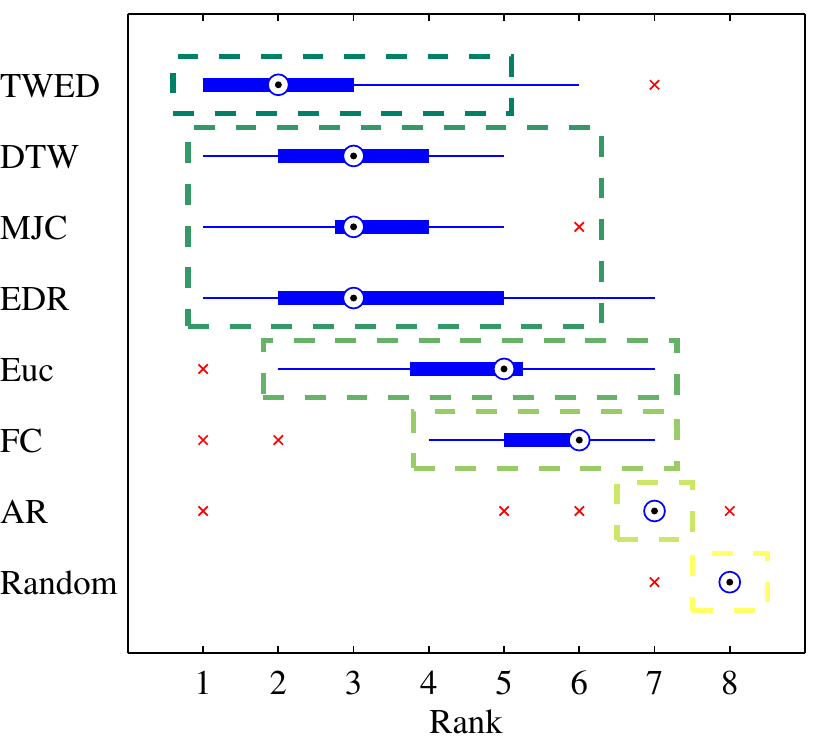}
	\end{center}
	\caption{Box plot for the distribution of performance ranks of each measure across data sets. The dashed lines denote statistically significantly equivalent groups of measures ($p<0.05$, Sec.~\ref{sec:Eval_StatSig}).}
	\label{fig:RankComp}
\end{figure}

\begin{figure}[t]
	\begin{center}
	\includegraphics{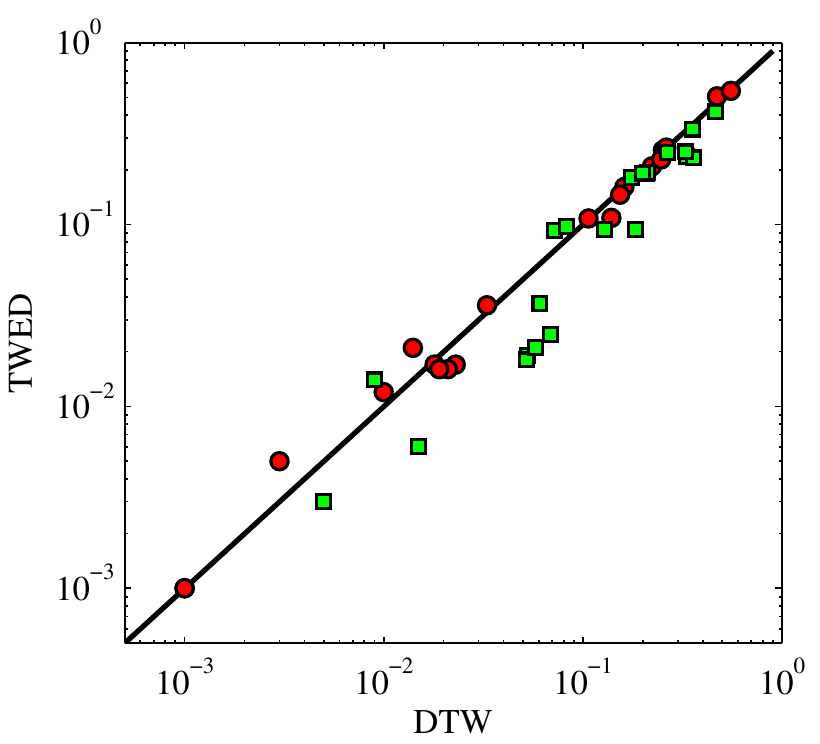}
	\end{center}
	\caption{Error ratios comparison between DTW and TWED (notice the logarithmic axes). The lower-right triangular part corresponds to TWED outperforming DTW, whereas the upper-left part corresponds to the opposite case. The green squares indicate statistically significant performance differences ($p<0.05$, Sec.~\ref{sec:Eval_StatSig}).}
	\label{fig:ErrComp}
\end{figure}

In general, we see that TWED outperforms the other measures in several data sets, with an average rank of 2.29 (Table~\ref{tab:Errors}). In fact, if we compare the considered measures on a more global scale, taking the matched error ratios across data sets (Sec.~\ref{sec:Eval_StatSig}), we obtain that TWED is statistically significantly superior to the rest (Fig.~\ref{fig:RankComp}). Next, we see that DTW, MJC, and EDR form a group of equivalent measures, with no statistically significant difference between them. The performed statistical analysis also separates the remaining measures from these and also between themselves. Apart from this more global analysis, further pairwise comparisons can be made, confirming the aforementioned global tendencies (Fig.~\ref{fig:ErrComp}).

\subsection{Classification performance: test vs.~train}
\label{sec:Results_Train}

\begin{figure*}[t]
	\begin{center}
	\includegraphics[width=0.49\linewidth]{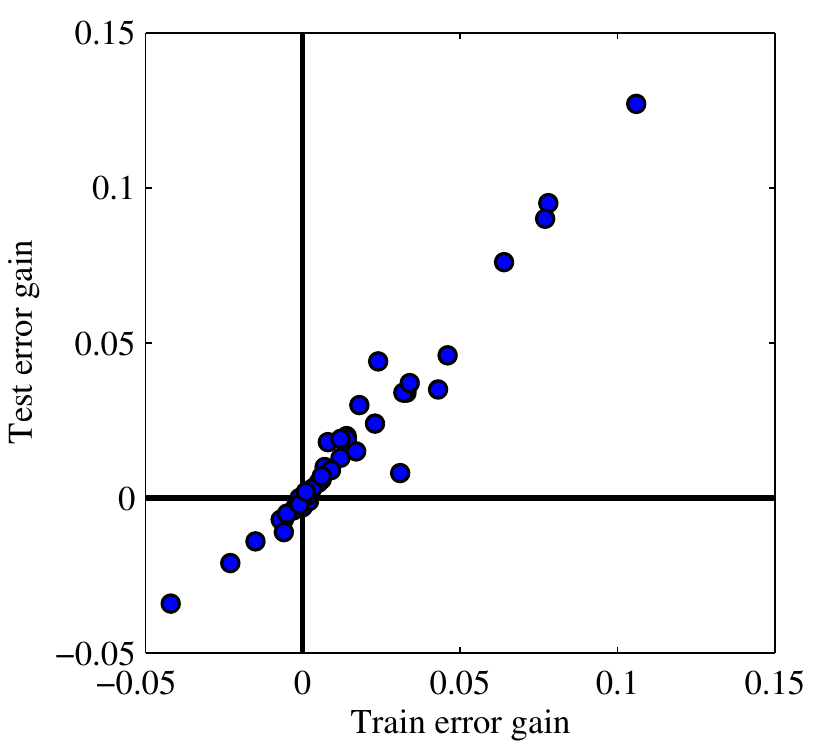}
	\includegraphics[width=0.49\linewidth]{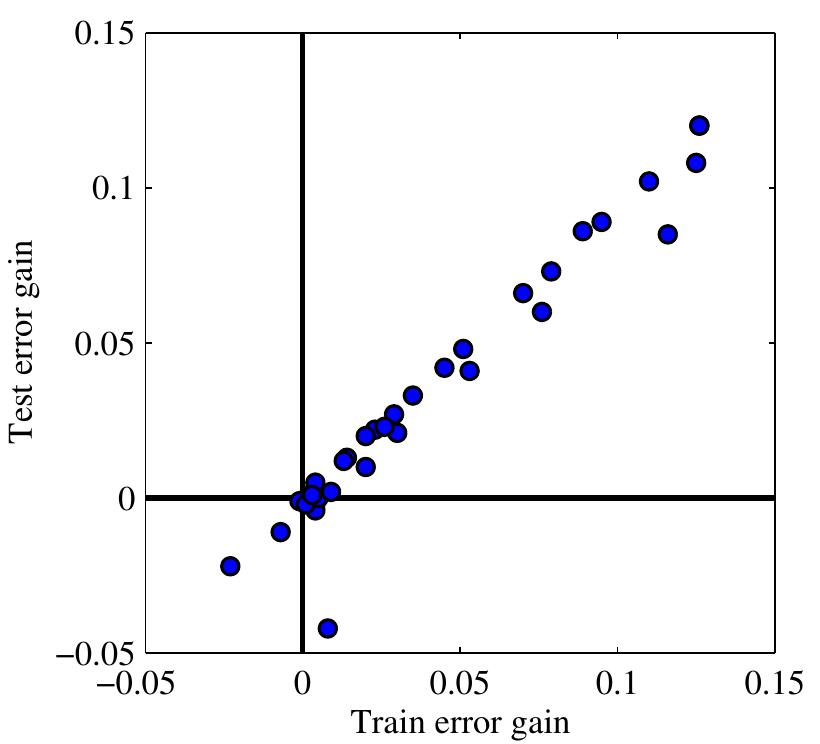}
	\end{center}
	\caption{Texas sharpshooter plots for TWED against DTW (left) and Euclidean distance (right). Here, error gain is measured by subtracting the TWED error ratio from the one of DTW/Euclidean. Dots around the diagonal indicate agreement of error gain for train and test. False positives, i.e.,~dots in the lower-right quadrant, indicate that TWED, being the best measure after training, does not reach the lowest error at testing. For instance, in the case of TWED vs.~Euclidean (right), the \textit{OliveOil} data set false positive stands out at coordinates $(0.008,-0.042)$ (see also Table~\ref{tab:TrainTestOutperf}). For further details on the construction of Texas sharpshooter plots we refer to~\citet{Batista11SDM}.}
	\label{fig:Texas}
\end{figure*}

For choosing the most optimal parameters for a given measure and data set we solely dispose of the training data. Hence, it is important to know whether the error ratios for training and testing sets are similar, otherwise one could be incurring into the so-called ``Texas sharpshooter fallacy''~\citep{Batista11SDM}, i.e.,~one could not predict a measure's utility ahead of time by just looking at training data. For comparing train and test error ratios, we can compute an error gain value for a couple of measures on each data set and check whether such values for train and test agree. To do so, a kind of real-valued contingency table can be plotted, called the ``Texas sharpshooter plot'' by~\citet{Batista11SDM}. Due to space reasons, we here only show such contingency tables for TWED against DTW and Euclidean distance (Fig.~\ref{fig:Texas}). The results show that error gains between TWED and DTW/Euclidean mostly agree between training and testing. As mentioned in Sec.~\ref{sec:Eval_CrossValid}, a full, raw account of train and test errors is available online. Having a close look at those full results, we can see that, in general, the best-performing measure at the training stage is also the best-performing measure at the testing stage. The few exceptions can be easily listed (Table~\ref{tab:TrainTestOutperf}). The relative rankings for the measures that do not perform best also mostly agree between train and test.

\begin{table}[t]
	\begin{center}
	\begin{tabular}{rl|ccc}
	\hline
	\hline
	\# 	& Data set				& Measure		& Outperf. by	& Gain \\
	\hline
	3 	& \textit{Beef}  		& FC				& EDR 				& 0.049 \\
	4 	& \textit{CBF} 			& TWED			& DTW 				& $<$0.001 \\
	7 	& \textit{Coffee}  		& DTW			& FC 				& 0.004 \\
	12 	& \textit{ECG200}  		& TWED			& MJC 				& 0.002 \\
	15	& \textit{FaceFour}		& MJC			& EDR				& 0.007 \\
	18	& \textit{Gun\_Point}	& EDR			& MJC				& 0.004 \\
	19 	& \textit{Haptics}  		& TWED			& MJC 				& 0.009 \\
	21 	& \textit{ItalyPowerDemand} & DTW		& EDR 				& 0.001 \\
	28 	& \textit{NonInvasiveFetalECG2} & Euclidean & TWED			& 0.001 \\
	29 	& \textit{OliveOil} 		& Euclidean 		& TWED				& 0.008 \\
	39 	& \textit{TwoLeadECG}	& TWED 			& DTW				& $<$0.001 \\
	\hline
	\hline
	\end{tabular}
	\end{center}
	\caption{List of best-performing measures in testing (the column ``Measure'') but actually outperformed by others in training (the column ``Outperf.~by''). The column ``Gain'' corresponds to the absolute value of the train error gain, i.e.,~the absolute difference between error ratios at training stage (see also Fig.~\ref{fig:Texas}).}
	\label{tab:TrainTestOutperf}
\end{table}

\subsection{Parameter assessment}
\label{sec:Results_Params}

\begin{figure*}[!t]
	\begin{center}
	\includegraphics[width=0.99\linewidth]{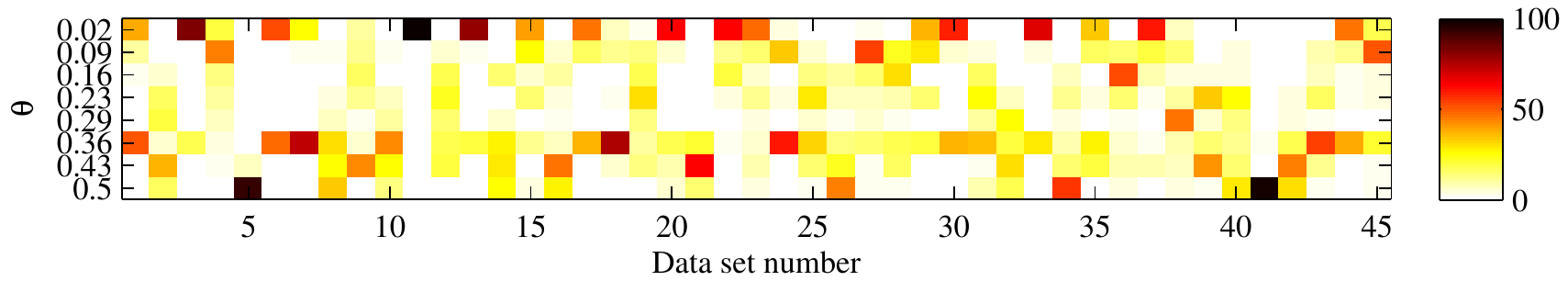}\\
	\includegraphics[width=0.99\linewidth]{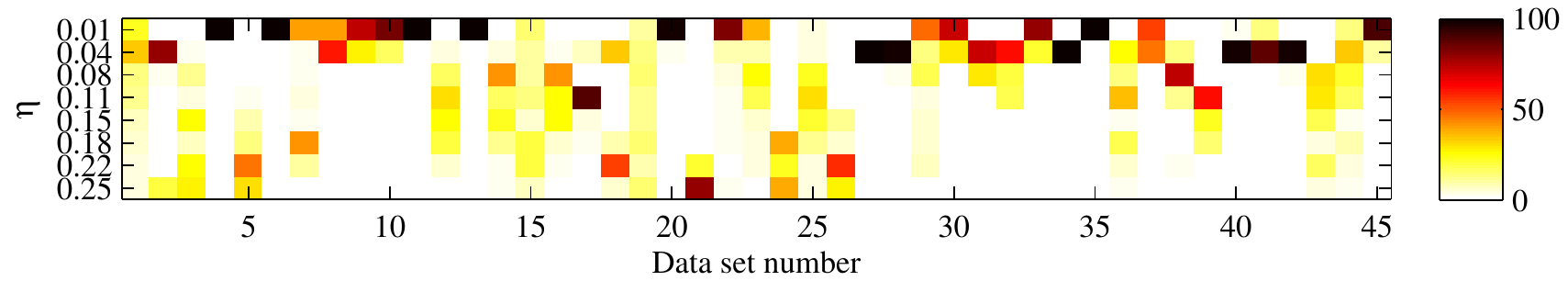}\\
	\includegraphics[width=0.99\linewidth]{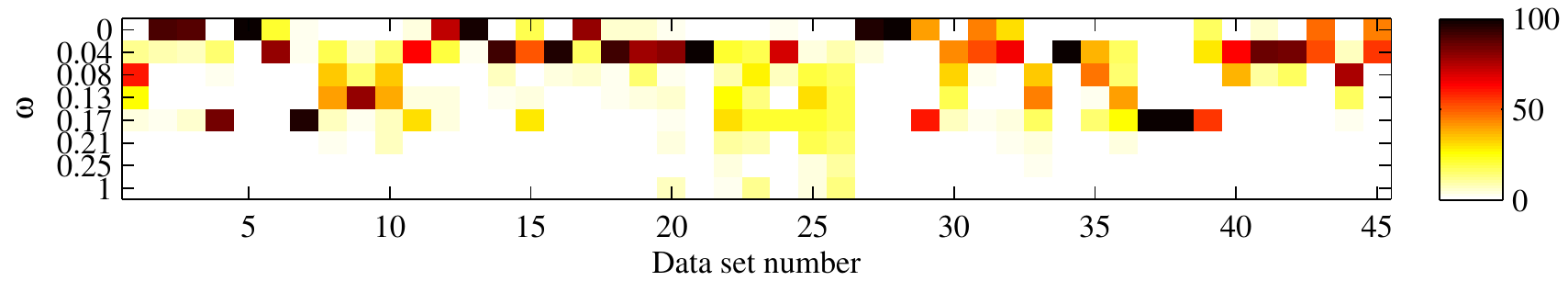}\\
	\includegraphics[width=0.99\linewidth]{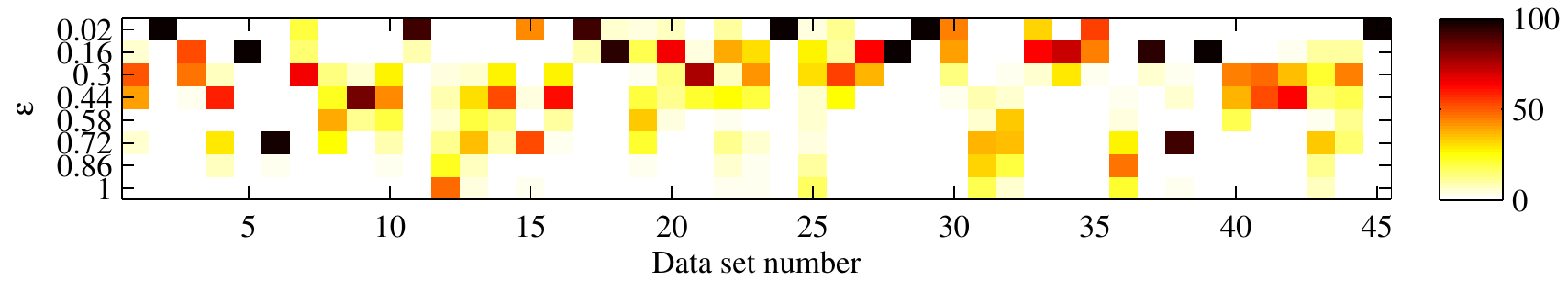}\\
	\includegraphics[width=0.99\linewidth]{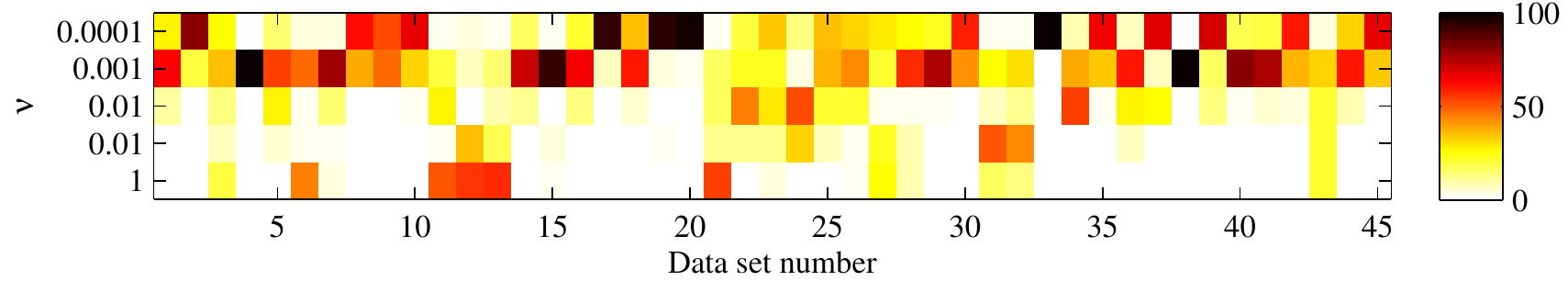}\\
	\includegraphics[width=0.99\linewidth]{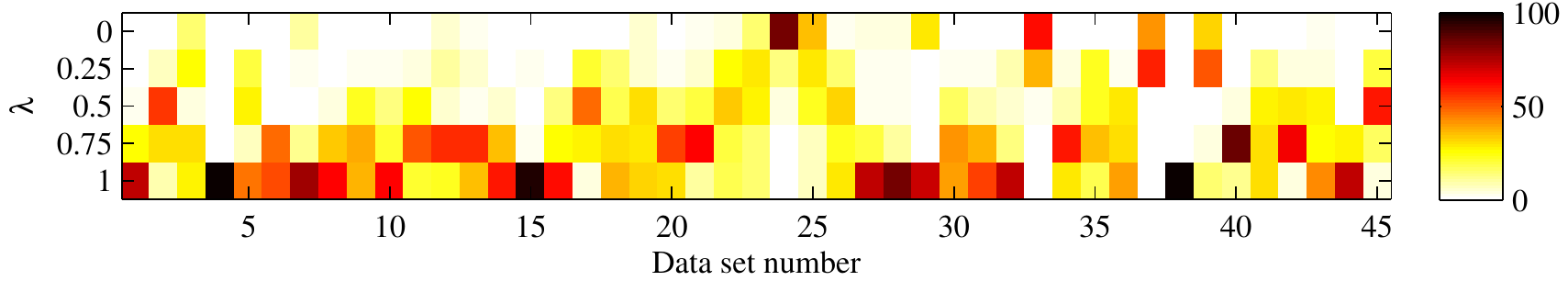}\\
	\includegraphics[width=0.99\linewidth]{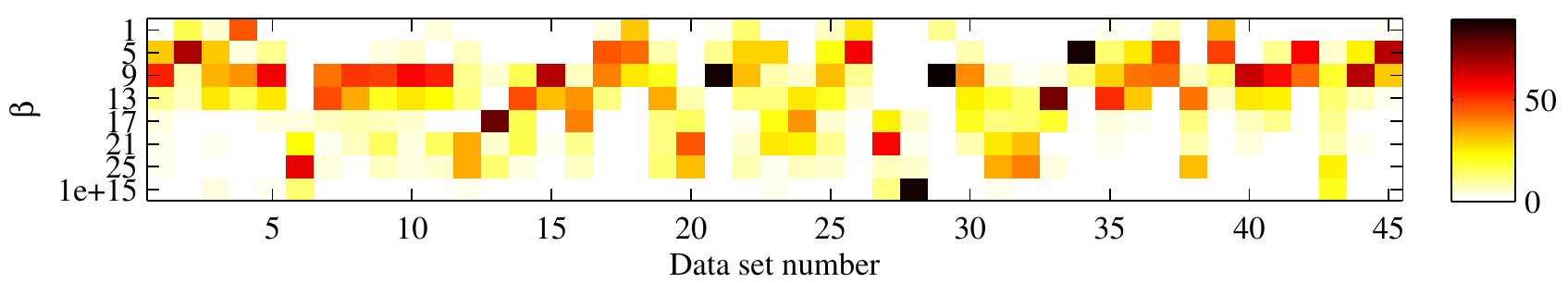}
	\end{center}
	\caption{Percentage of times (color code) that a given parameter value (vertical axis) is chosen for each data set (horizontal axis; for the names behind each number see Table~\ref{tab:Errors}). From top to bottom, the plots correspond to FC ($\theta$), AR ($\eta$), DTW ($\omega$), EDR ($\varepsilon$), TWED ($\nu$), TWED ($\lambda$), and MJC ($\beta$).}
	\label{fig:Params}
\end{figure*}

We finally report on the parameters chosen for each measure after training with 66\% of balanced data (Fig.~\ref{fig:Params}). Firstly, we observe that, in the vast majority of cases, a specific value for a given parameter is consistently chosen across the $20\times 3$ performed iterations (we see clear peaks in the distributions of Fig.~\ref{fig:Params}). Among these consistent choices, perhaps TWED and MJC present the most spread distributions. Such aspect, together with the fairly good accuracies obtained for these two specific measures (Sec.~\ref{sec:Results_Test}), indicates a certain degree of robustness against specific parameter choices. This is a very desirable quality of a time series similarity measure, even more if we have to train a classifier with a potentially incomplete set of training instances.

Next, we see that the selected parameters are generally not in the borders of the specified ranges, thus indicating that a reasonable choice has been made (Fig~\ref{fig:Params}). This is particularly true for DTW and EDR. The only measure that could potentially benefit from reconsidering the parameters' range is TWED. As it can be seen, $\nu$ and $\lambda$ seem to be consistently chosen in the lower and upper parts of the specified ranges, respectively. This suggests that the best combination for some data sets could lie outside the parameter space outlined by~\citet{Marteau09TPAMI}, i.e., in $0<\nu<10^{-4}$ and/or $\lambda>1$. If that was the case, TWED could potentially achieve even much higher accuracies. Interestingly, TWED is not the best-performing measure for some of the data sets where `border' parameter values are chosen (e.g.,~\textit{CBF}, \textit{Fish}, \textit{StarLightCurves}, \textit{TwoPatterns}).

Finally, we can comment on the particularities of some data sets with relation to classification. For instance, we see that a relatively large window parameter $\omega$ (DTW) is chosen for data sets 36 to 39 (i.e.,~\textit{Synthetic\_control}, \textit{Trace}, \textit{Two\_Patterns}, and \textit{TwoLeadECG}). This denotes that tracking alignments or warping paths beyond the main diagonal of $D$ (Eq.~\ref{eq:DTW}) might be advantageous for classification in these data sets. Interestingly, the stiffness parameter $\nu$ (TWED), which accounts for a similar but opposite concept (Sec.~\ref{sec:TWED}), takes relatively small values. Such agreement across different measures reinforces the hypothesis that tracking intricate alignments or strongly warped paths may be advantageous for these data sets. Analogous and complementary conclusions can be derived for other data sets. For instance, in data sets 11 (\textit{DiatomSizeReduction}) and 13 (\textit{ECGFiveDays}), a small number of both FCs $\theta$ and AR coefficients $\eta$ is chosen. As FC and AR achieve competitive accuracies in those specific data sets, we could suspect that low-frequency components are important for correctly classifying the instances in those data sets (Secs.~\ref{sec:FC} and~\ref{sec:AR}). 

\section{Conclusion}
\label{sec:Conclu}

From a general perspective, the obtained results show that there is a group of equivalent similarity measures, with no statistically significant differences among them (DTW, EDR, and MJC). The existing literature suggests that some longest common sub-sequence approaches~\citep{Gusfield97BOOK}, together with alternative variants of DTW and EDR~\cite[e.g.,][]{Sakoe78TASLP,Morse07SIGMOD}, could potentially join this group~\citep{Marteau09TPAMI,Wang12DMKD}. However, according to the results reported here, the TWED measure originally proposed by~\citet{Marteau09TPAMI} seems to consistently outperform all the considered distances, including DTW, EDR, and MJC. Thus, we believe this often unconsidered measure should take a baseline role in future evaluations of time series similarity measures (beyond accuracy, additional properties enumerated in Sec.~\ref{sec:TWED} make it also very attractive). The Euclidean distance, although somehow competitive, generally performs statistically significantly worse than TWED, DTW, MJC, and EDR. Its accuracy on large data sets was also not very impressive. Below Euclidean distance, but statistically significantly above the random baseline, we find FC and AR measures. Of course, the general statements above do not exclude the possibility that a particular measure or variant could be very well-suited for a specific data set and statistically significantly outperform the rest~\cite[cf.][]{Keogh03DMKD}. In Sec.~\ref{sec:Results_Test} have enumerated several examples of that.

When comparing train and test errors, we have seen that these mostly agree, with train errors generally providing a good guess of the test errors on unseen data. We have listed some notable exceptions to this rule and used Texas sharpshooter plots to further assess this aspect for TWED vs.~DTW and Euclidean. When assessing the best parameter choices for each measure, we have seen that the considered ranges are typically suitable for the task at hand. We have also discussed some particularities regarding parameter choices and the nature of a few data sets.

The similarity measure is a crucial step in computational approaches dealing with time series. However, there are some additional issues worth mentioning, in particular with regard to post-processing steps focused on improving similarity assessments (pre-processing steps are sufficiently well-discussed in the existing literature~\citep[see, e.g.,][and references therein]{Keogh03DMKD,Han05BOOK,Wang12DMKD}. A very interesting post-processing step is the complexity-invariant correction factor introduced by~\citet{Batista11SDM}. Such correction factor prevents from assigning low dissimilarity values to time series of different complexity, thus preventing the inclusion of time series of different nature in the same cluster. The way to assess complexity depends on the situation, but~\citet{Batista11SDM} introduce a quite straightforward way: the L$_2$ norm of the sample-based derivative of a time series. Overall, considering different types of `invariance' is a sensible approach~\cite[][provide a good overview]{Batista11SDM}. Here, we have already implicitly considered a number of them, although more as a pre-processing or method-specific strategy: global amplitude and scale invariance (z-normalization), warping invariance (any elastic measure, in our case DTW, EDR, TWED, and MJC), phase invariance (AR\footnote{For FC we use both phase and magnitude (Sec.~\ref{sec:FC}).}), and occlusion invariance (EDR and TWED).

Another interesting post-processing step is the hubness correction for time series classification introduced by~\citet{Radovanovic10SDM}. Based on the finding that some instances in high-dimensional spaces tend to become hubs by being unexpectedly (and usually wrongly) considered nearest neighbors of several other instances, a correction factor can be introduced. This usually does not harm classification accuracy and can definitely improve performance for some data sets~\citep{Radovanovic10SDM}. A further strategy for enhancing time series similarity and potentially reducing hubness is the use of unsupervised clustering algorithms to prune nearest neighbor candidates~\citep{Serra12PRL}. Future work should focus on the real quantitative impact of strategies for enhancing time series similarity like the ones above, with a special emphasis on its impact to different measures and classification schemes.

The empirical comparison of multiple approaches across a large-scale case basis is an important and necessary step towards any mature research field. Besides getting a more global picture and highlighting relevant approaches, it pushes towards unified validation procedures and analysis tools. It is hoped that this article will serve as a steppingstone for those interested in advancing in time series similarity, clustering, and classification.

\section*{Acknowledgements}

We thank the people who made available or contributed to the UCR time series repository. This research has been funded by 2009-SGR-1434 from Generalitat de Catalunya, JAEDOC069/2010 from Consejo Superior de Investigaciones Cient\'ificas, and TIN2009-13692-C03-01 and TIN2012-38450-C03-03 from the Spanish Government.

\bibliographystyle{elsarticle-harv}
\bibliography{bibjserra}

\end{document}